\documentclass[letterpaper, 10 pt, conference]{ieeeconf}  

\IEEEoverridecommandlockouts                              
\usepackage{graphicx}   
\usepackage{subcaption}
\usepackage{amssymb}
\usepackage{hyperref}
\usepackage{amsmath}
\usepackage{algorithm}
\usepackage{mathtools}  
\usepackage{xcolor}
\usepackage{booktabs}

\let\labelindent\IEEElabelindent
\usepackage{enumitem}

\newcommand{\kmax}{\mathop{\mathrm{kmax}}}
\newcommand{\grpkmax}{\mathop{\mathrm{grpkmax}}}

\newcommand{\sgn}{\mathop{\mathrm{sgn}}}

\newtheorem{theorem}{Theorem}

\newcommand{\lasso}{Lasso}
\newcommand{\glasso}{GroupLasso}
\newcommand{\sglasso}{SparseGroupLasso}
\newcommand{\gkmax}{GroupKmax}

\title{\LARGE \bf
Sparsity via Sparse Group $k$-max Regularization
}

\author{Qinghua Tao, Xiangming Xi, Jun Xu and Johan A.K. Suykens
\thanks{Qinghua Tao and Johan A.K. Suykens are with STADIUS, ESAT, KU Leuven, Belgium
        {\tt\small qinghua.tao, johan.suykens@esat.kuleuven.be}}%
\thanks{Xiangming Xi is with Research Center for Intelligent Robotics, Zhejiang Lab, Hangzhou, Zhejiang, China  {\tt\small xixiangming@gmail.com}}%
\thanks{Jun Xu is with School of Mechanical Engineering and Automation, Harbin Institute of Technology, Shenzhen, 518055, China,  {\tt\small xujunqgy@hit.edu.cn}}%
\thanks{Qinghua Tao and Xiangming Xi contribute equally, and Jun Xu is the corresponding author.}
\thanks{Xiangming Xi acknowledges National Natural Science Foundation of China (Grant No. U22A6001). 
Johan A.K. Suykens and Qinghua Tao acknowledge the support of ERC Advanced Grant E-DUALITY (787960), KU Leuven Grant CoE PFV/10/002, and Grant  FWO G0A4917N, EU H2020 ICT-48 Network TAILOR, and Leuven.AI Institute. Jun Xu acknowledges the National Natural Science Foundation of China (U1813224, 62173113), and Natural Science Foundation of Guangdong Province of China (2022A1515011584). }
}

\begin{document}

\maketitle
\thispagestyle{empty}
\pagestyle{empty}

\begin{abstract}


    For the linear inverse problem with sparsity constraints, the $l_0$ regularized problem is NP-hard, and existing approaches either utilize greedy algorithms to find almost-optimal solutions or to approximate the $l_0$ regularization with its convex counterparts. 
    In this paper, we propose a novel and concise regularization, namely the sparse group $k$-max regularization, which can not only simultaneously enhance the group-wise and in-group sparsity, but also 
    casts no additional restraints on the magnitude of variables in each group, which is especially important for variables at different scales, 
    so that it approximate the $l_0$ norm more closely. We also establish an iterative soft thresholding algorithm with local optimality conditions and complexity analysis provided. Through numerical experiments on both synthetic and real-world datasets, we verify the effectiveness and flexibility of the proposed method.

\end{abstract}

\addtolength{\textheight}{-3cm}   

\section{INTRODUCTION}
    In general, the inputs and outputs of a system are observable while the system parameters are unknown especially for much more complicated large systems. A linear inverse problem (LIP) is to infer such unknown parameters from the observations of a linear system, i.e.,
    \begin{equation}\label{eqn:linear_system}
        \boldsymbol a^T \boldsymbol \phi + \epsilon = y,
    \end{equation}
    where $\boldsymbol \phi \in \mathcal{X}$ and $y \in \mathcal{Y}$ are observed system inputs and outputs, $\boldsymbol a \in \mathbb{R}^n$ is the system parameter to be identified, and $\epsilon \sim \mathcal{N}(0, \sigma^2)$ is the white noise with variance $\sigma^2$. 

    LIPs have been successfully applied in many real world engineering problems, such as compressed sensing \cite{Blumensath2009}, image processing \cite{Gilton2020}, model selection \cite{Yuan2006}, and so on.
    Considering that variables in (\ref{eqn:linear_system}) might have different dimensionalities in different applications, we take the following formulation of LIPs with vectorized variables for simplicity \cite{Daubechies2004},    
    \begin{equation}\label{eqn:LIP-general}
        \sum_{i = 1}^m \Phi_i \boldsymbol{x}_i + \boldsymbol \epsilon = \boldsymbol y,
    \end{equation}
    where $\Phi_i \in \mathbb{R}^{n \times d_i}$ ($\forall i \in [m] \triangleq \{ 1, \cdots, m\}$, and $\boldsymbol y \in \mathbb{R}^{n}$ are observations, $\boldsymbol{x}_i \in \mathbb{R}^{d_i}$ ($\forall i \in [m]$) are variables to be identified, $\boldsymbol \epsilon \in \mathbb{R}^n$ is the white noise with each two entries being independent.

    In order to identify the variables in (\ref{eqn:LIP-general}), it is common to solve the following optimization problem based on the least square loss function, i.e., 
    \begin{equation}\label{prob:no-penalty}
        \min_{\boldsymbol x} \quad \frac{1}{2n}||\boldsymbol y - \sum_{i = 1}^m \Phi_i \boldsymbol{x}_i||_2^2.
    \end{equation}
    In practice, it is difficult to solve (\ref{prob:no-penalty}) despite of its convexity since $\Phi_i$ are generally ill-conditioned or underdetermined ($n \ll \sum_{i \in [m]} d_i$) \cite{Simon2013a}.

    In applications such as model selection and compressed sensing, other properties of the solutions are desired, among which the sparsity is of more concern. A trivial sparsity metric of a vector $\boldsymbol{x} \in \mathbb{R}^n$ is the $l_0$ pseudo-norm (abbr. $l_0$ norm), denoted as $|| \boldsymbol{x} ||_0$, which counts the number of its nonzero entries. Consequently,  adding $l_0$ regularization to (\ref{prob:no-penalty}) leads to the following unconstrained problem,
    \begin{equation}\label{prob:l0-min}
        \min_{\boldsymbol x} \quad \frac{1}{2n}||\boldsymbol y - \sum_{i = 1}^m \Phi_i \boldsymbol{x}_i||_2^2 + \lambda \sum_{i = 1}^m ||\boldsymbol{x}_i||_0,
    \end{equation}
    where $\lambda > 0$ is a regularization parameter.

    To solve (\ref{prob:l0-min}), researchers need to address the discontinuity and non-convexity induced by the $l_0$ norm, and various greedy algorithms have been proposed, such as the orthogonal matching pursuit  \cite{Tropp2007} and subspace pursuit algorithm \cite{Dai2009}. 
    
    Though these greedy algorithms generally achieve comparative performances, their solutions might not be guaranteed to be optimal \cite{Zhang2015b} and the NP-hardness of (\ref{prob:l0-min}) requires more computational resources for large-scale problems \cite{Natarajan1995}.

    In order to guarantee the optimality of the solutions to (\ref{prob:l0-min}), researchers propose to approximate the $l_0$ norm with its convex counterparts. As its tightest  convex envelope, the $l_1$ norm attracts the most attention \cite{Tibshirani1996}, and the resulted problem can be expressed as 	
    \begin{equation}\label{prob:l1-norm}
	   \min_{\boldsymbol x} \quad \frac{1}{2}||\boldsymbol y -\sum_{i = 1}^m  {\Phi} \boldsymbol{x}_i||_2^2+ \lambda \sum_{i = 1}^m ||\boldsymbol{x}_i||_1.
    \end{equation}
    It is indeed the famous lasso (least absolute shrinkage and selection operator) in the community. Since (\ref{prob:l1-norm}) is convex, it can be efficiently and effectively solved  using the coordinate descent \cite{Tibshirani1996}, the alternating direction method of multipliers (ADMM) \cite{Boyd2011}, etc. Beside the $l_1$ norm, researchers also propose the $l_p$ norm ($1 < p < 2$) \cite{Donoho2011}, $l_2$ norm (the ridge regression) \cite{Hoerl1970}, $k$-support norm \cite{Argyriou2012}, smoothly clipped absolute deviation regularization \cite{Fan2001}, or the combination of different relaxations \cite{Zou2005}. 
    
    Despite of the advantages of the convex relaxation techniques, there are still some challenges to be addressed. First, as $p > 1$ grows larger, the solutions obtained using the $l_p$ norm might be less sparse \cite{Zhang2015b}. Second, some relaxations of the $l_0$ norm regularize all entries of the variable in their magnitudes  \cite{Argyriou2012}, which behaves differently from the $l_0$ norm. In order to tackle these challenges, Cand\`{e}s et al. propose the reweighted $l_1$ norm to adaptively regularize different variable entries according to their magnitudes \cite{Candes2008},  and Huang et al. propose the two-level $l_1$ norm, which partitions the variable entries into two groups according to their order and assigns different regularization parameters \cite{Huang2015}.

    Beside the overall sparsity of the solution, which treats $\boldsymbol{x}_i$ ($\forall i \in [m]$) in (\ref{prob:l0-min}) equally, some applications pay more attention on the grouping characteristic of the variables. For example, in gene expressions, each $\boldsymbol{x}_i$ may represent a gene pathway, and different pathways function differently \cite{Yuan2006}. Therefore, these variables need different treatments. One strategy is to treat entries in a variable ($\boldsymbol{x}_i$)  as a whole and to investigate their influences on the observations. Another is an extension of the first strategy that the individual behavior of entries in a group should also be taken into consideration. In the literature, a representative of the the first strategy is the group lasso \cite{Yuan2006}, which takes the following expression,  
    \begin{equation}\label{prob:grplasso}
		\min_{\boldsymbol x} \quad \frac{1}{2}||\boldsymbol y - \sum_{i = 1}^m \Phi_i \boldsymbol{x}_i||_2^2+ \lambda \sum_{i = 1}^m ||\boldsymbol{x}_i||_K,
	\end{equation}
	where $||\boldsymbol{x}_i||_K = (\boldsymbol{x}_i^T K \boldsymbol \theta_i)^{1/2}$ and $K$ is a positive definite matrix. Other examples include  the generalization for infinite-dimensional setting \cite{Bach2008} and the joint variable selection  for multi-task learning \cite{Obozinski2010}. The representative of the second strategy is the sparse group lasso \cite{Simon2013a}, where an extra $l_1$ regularization is added to the group lasso to enhance sparsity within each group, and the expression is as follows,
	\begin{equation}\label{prob:spgrplasso}
		\min_{\boldsymbol x} \quad \frac{1}{2}||\boldsymbol y-\sum_{i = 1}^m \Phi_i \boldsymbol{x}_i||_2^2+ \lambda \sum_{i = 1}^m ||\boldsymbol{x}_i||_K + \mu \sum_{i = 1}^m ||\boldsymbol{x}_i||_1,
	\end{equation}
	where $\mu > 0$ is the in-group regularization parameter. For problem  (\ref{prob:grplasso}) and (\ref{prob:spgrplasso}), existing optimization approaches include the block coordinate descent algorithm \cite{Yuan2006} and  the proximal point algorithm \cite{Yang2013}.

	

    In summary, the single or joint considerations on group-wise and in-group sparsity constraints for LIPs can be obtained either with the non-convex and discontinuous $l_0$ norm or with its convex relaxations, and much progress has been make. However, there are still challenges to be tackled. 
    The existing works mainly  focus on either the enhanced sparsity in a single group of variables or the group-structured sparsity. It can be seen that there exist different penalties with enhanced sparsity, but these penalties have not been incorporated under the group-structured sparsity framework to achieve enhanced sparsity  both within and across groups. 
    In this paper, to  enhance the group-wise and in-group sparsity of the solutions simultaneously, we introduce the sparse group $k$-max regularization, which approximate the $l_0$ norm more closely by penalizing  a portion of the grouped variables while casting no additional restraints on the others (Section \ref{sec:groupkmax}). 
        We also propose an iterative soft thresholding (IST) algorithm for  the regularized problem with the proposed regularization and prove the local optimality conditions and complexity analysis (Section \ref{sec:alg}). Numerical experiments verify that the proposed group $k$-max regularization can flexibly achieve enhanced sparsity while maintaining accuracy (Section \ref{sec:exp}).

\section{SPARSE GROUP $k$-MAX REGULARIZATION}\label{sec:groupkmax}

    In this section, we introduce a novel relaxation of the $l_0$ norm. Despite of its non-convexity, it can not only enhance the group-wise and the in-group sparsity, but also remove restraints on the magnitudes of nonzero variables. 

    Since the $l_0$ norm evaluates the number of nonzero entries in $\boldsymbol x$, an intuitive evaluation is to sort the entries in the descending order, and to localize the first  entry with zero value with entry index $(k+1)$, which indicates that the sparsity of $\boldsymbol{x}$ is $k$ (if $\boldsymbol x$ has a zero entry). In this way, (\ref{prob:l0-min})  can be approximated by the following \textit{sparse group $k$-max regularized problem}, i.e.,
	\begin{equation}\label{prob:group_kmax_optimization}
		\min_{\boldsymbol x} \  \frac{1}{2}|| \boldsymbol y-\sum_{i = 1}^m \Phi_i \boldsymbol{x}_i ||_{2}^2 + \lambda \sum_{i = 1}^m L_{\grpkmax}^{k_i} (\boldsymbol{x}_i),
	\end{equation}
    where $L_{\grpkmax}^{k_i} (\cdot)$ is the sparse group $k$-max regularization corresponding to $\boldsymbol{x}_i$ and is formulated as follows, 
	\begin{equation}\label{eqn:group_kmax}
			L_{\grpkmax}^{k_i} (\boldsymbol{x}_i) = \sum_{j\in I_{k_i}^{\leq}(\boldsymbol x_i)} |x_i(j)|,
	\end{equation}
    where $x_i(j)$ represents the $j$-th entry of $\boldsymbol{x}_i$. In (\ref{eqn:group_kmax}), the core component is the index set $I_{k_i}^{\leq}(\boldsymbol{x}_i)$ which determines which entries should be included in the regularization. In order to obtain the index set properly, we denote the $k_i$-th maximal absolute value of $\boldsymbol{x}_i$ as $t_{ik_i}$, and thus, the index set can be defined as follows,
	\begin{equation}\label{kmax_index}
		\begin{array}{ll}
			& I_{k_i}(\boldsymbol{x}_i) \triangleq \{i \in [d_i]: |x(i)| = t_{ik_i}\},\\
			& I_{k_i}^+(\boldsymbol{x}_i) \triangleq \{i \in [d_i]: |x(i)| > t_{ik_i}\},\\
			& I_{k_i}^-(\boldsymbol{x}_i) \triangleq \{i \in [d_i]: |x(i)| < t_{ik_i}\}, \\
			& I_{k_i}^{\leq}(\boldsymbol{x}_i) \triangleq \{i \in [d_i]: |x(i)| \leq t_{ik_i}\}=I_{k_i}(\boldsymbol{x}_i)\bigcup I_{k_i}^-(\boldsymbol{x}_i),\\
		\end{array}
	\end{equation}
    It is obvious that $|I_k(\boldsymbol x)| + |I_k^+(\boldsymbol x)| + |I_k^-(\boldsymbol x)| = d$, where  $|\cdot|$ stands for the cardinality operator of a countable set.


    Based on (\ref{eqn:group_kmax}), we can make a comparison with other relaxations of the $l_0$ norm. 
    \begin{enumerate}[label=(\arabic*),itemindent=15pt,labelsep=3pt,fullwidth,parsep=0pt,labelwidth=0pt]
        \item Different from the $l_p$ ($p \geq 1$) norm, which are generally convex, the proposed regularization is indeed concave. Besides, the proposed regularization is continuous but non-smooth. This somehow increases the complexity of the optimization algorithms, and it can not guarantee that any local optimum is globally optimal.

        \item The proposed regularization emphasizes the sparsity within the group of variables which shows overall non-sparsity, which in turn simultaneously enhances the group-wise and in-group sparsity, which is more concise than the sparse group lasso \cite{Simon2013a}. 
        
        \item The proposed regularization only regularizes the smallest $(d_i - k_i)$ entries while the larger ones are absent in the expression. This benefits especially in the case that entries in $\boldsymbol{x}_i$ have different scales and in-group sparsity is desired. This property makes it to approach the $l_0$ norm more closely and distinguishable from some of the $l_p$ regularizations which treat all entries equally.

        \item The proposed regularization requires the ordering of entries of $\boldsymbol{x}_i$ which might be time consuming, which is also used in the two-level $l_1$ regularization \cite{Huang2015}.

        \item When $k_i = 0$, $\forall i$, the proposed regularization degrades to the $l_1$ norm.
    \end{enumerate}

	Therefore, the proposed sparse group $k$-max regularization is  an analogy with the $l_0$ norm bringing enhanced group-wise and in-group sparsity.


	

\section{ALGORITHMS}\label{sec:alg}
    In this section, we first introduce the iterative soft thresholding algorithm for the sparse group $k$-max regularized problem (\ref{prob:group_kmax_optimization}), and then provide theoretical analysis with respect to the optimality conditions, the convergence analysis, as well as brief complexity analysis. 
    

\subsection{Modified Iterative Soft Thresholding Algorithm}
    Iterative soft thresholding algorithm is a classical algorithm for optimization problems with sparsity constraints \cite{Liu2019h}. 
    In the context related to linear inverse problems, the existing iterative soft thresholding algorithms mainly focus on the convex relaxation versions \cite{Daubechies2004}. Recently, with the non-convex relaxation of LIPs emerging, IST algorithms have also been developed with theoretical guarantee in convergence and optimality. An example is the two-level soft thresholding algorithm proposed for the two-level $l_1$ norm regularized problem \cite{Huang2015}. 

    In order to solve the sparse group $k$-max regularized problem (\ref{prob:group_kmax_optimization}), which is neither convex nor differentiable, we derive the group $k$-max soft shrinkage operator. We denote the operator for $\boldsymbol{x}_i$,  $\forall i \in [m]$, as $\mathbb{S}_{\grpkmax}^{ k_i, \lambda}(\boldsymbol{x}_i) \in \mathbb{R}^n$, $\forall j \in [d_i]$, i.e.,
	\begin{equation}\label{soft_new}
            \begin{array}{lll}
                 & \mathbb{S}_{\grpkmax}^{ k_i, \lambda}(\boldsymbol{x}_i) (j)   \\
                 = & \begin{cases}
			 x_{i}(j) - \lambda \sgn(x_{i}(j)), & \text{if} \ \ j\in I_{k_i}^{\leq}(\boldsymbol{x}_i), |x_{i}(j)| \geq \lambda,\\
			0, & \text{if} \ \ j \in I_{k_i}^{\leq}(\boldsymbol{x}_i), |x_{i}(j)| <  \lambda, \\
			x_{i}(j), & \text{if} \ \ j \in I_{k_i}^+(\boldsymbol{x}_i).  \\
		\end{cases}
            \end{array}
	\end{equation}
    where $\sgn(\cdot)$ is the sign function. As a result, the stationary condition of  (\ref{prob:group_kmax_optimization}) can be defined as follows, and its optimality condition will be explained in the next part.
	\begin{equation}\label{eq_stationary_group}
		\boldsymbol{x}_i = \mathbb{S}_{\grpkmax}^{ k_i, \lambda}(\boldsymbol{x}_i + \Phi_i^T(\boldsymbol y -\sum_{i=1}^m\Phi_i \boldsymbol{x}_i)), \forall i \in [m].
	\end{equation}
    
    Therefore, the modified IST algorithm is given in Algorithm \ref{alg:grpkmax}, where the algorithm terminates when the solutions stop decreasing (with precision $\delta$) or it reaches the maximal iteration number $T$.
    
    \begin{algorithm}[!htp]
		\caption{Group $k$-max Soft Thresholding Algorithm }\label{alg:grpkmax}
		\textbf{Input:} Given $\boldsymbol y$, $\Phi_i$ with $k_i$, $\forall i \in [m]$, $\lambda >0, \delta > 0$ , $T \in \mathbb Z_{++}$;\\
		\textbf{Output:} The estimator $\hat{\boldsymbol{x}_i}$,  $\forall i \in [m]$. \\
		$\bullet$ Set $t:=0$ and $\boldsymbol{x}_i^{(0)} := \Phi_i^T \boldsymbol y$;\\
		\textbf{Repeat}
		\begin{itemize}
			\item $\boldsymbol x_i^{(t+1)} = \mathbb{S}_{\grpkmax}^{ k_i, \lambda}(\boldsymbol{x}^{(t)}_i + \Phi_i^T(\boldsymbol y - \Phi_i \boldsymbol{x}^{(t)}_i)) $;
			\item Set $t:=t+1$.
		\end{itemize}
		\textbf{Until} $\sum_{i\in[m]}||\boldsymbol {x}^{(t)}_i - \boldsymbol{x}^{(t-1)}_i||_2 \leq \delta$ or $t\geq T$.\\
		$\bullet$ $\hat{\boldsymbol{x}_i} = \boldsymbol{x}^{(t)}_i$.\\
	\end{algorithm}

\subsection{Optimality Analysis}


The stationary condition (\ref{eq_stationary_group}) can be derived following Theorem 3.1 in \cite{Daubechies2004} and Theorem 2 in \cite{Huang2015} with some modifications.
As a basis, we first prove the simplest situration where $m = 1$, i.e., there is only one group of variables, and  (\ref{prob:group_kmax_optimization}) is reduced to 
\begin{equation}\label{prob:group_kmax_optimization_simple}
		\min_{\boldsymbol x} \  \frac{1}{2}|| \boldsymbol y- \Phi \boldsymbol{x} ||_{2}^2 + \lambda L_{\grpkmax}^{k} (\boldsymbol{x}),
	\end{equation}

For (\ref{prob:group_kmax_optimization_simple}), we have the following conclusions.

\begin{theorem}\label{localoptimal}
		If  $\exists  \boldsymbol  x^* \in \mathbb R^d$, for all possible $I_k^{\leq}(\boldsymbol u^*)$, where $\boldsymbol u^* = \boldsymbol x^*+\Phi^T(\boldsymbol{y}-\Phi \boldsymbol x^*)$, such that
		\begin{align}
			&\boldsymbol x^* = \mathbb{S}_{\kmax}^{k, \lambda}(\boldsymbol u^*),\label{eq_localoptimal}\\
			&t_k^+ >t_k + \lambda, \label{eq_localoptimal_add}
		\end{align}
		where  $t_k$ is the $k$-th maximal absolute value of $\boldsymbol u^*$  and $t_k^+ = \min_{j\in I_k^+(\boldsymbol u^*)}\{\boldsymbol u^*(j)\}$, then $\boldsymbol x^*$ is locally optimal to  (\ref{prob:group_kmax_optimization_simple}).
	\end{theorem}

	\begin{proof}
		Given a feasible solution $\boldsymbol x^*$ to (\ref{prob:group_kmax_optimization_simple}), we can compute $\boldsymbol u^* = x^*+\Phi^T(\boldsymbol y-\Phi x^*)$ and get the indices $I_{k}^{\leq}(\boldsymbol u^*)$ and $I_k^{+}(\boldsymbol u^*)$, which are briefly noted as
		\begin{align*}
			& I_k^{\leq} = \{i |\boldsymbol u^*(i)| \leq t_k\}, I_k^+ = \{j |\boldsymbol u^*(j)| > t_k\},
		\end{align*}
		where $t_k$ is the $k$-th maximal absolute value of $\boldsymbol u^*$. One can always find a small neighborhood of $\boldsymbol x^*$, say ${U}(\boldsymbol x^*, \delta), \delta >0$, such that for any $\boldsymbol x\in {U}(\boldsymbol x^*, \delta) $, there exists
		\begin{align*}
			&|(\boldsymbol x+\Phi^T(\boldsymbol y-\Phi \boldsymbol x))(i)|<|(\boldsymbol x+\Phi^T(\boldsymbol y-\Phi \boldsymbol x))(j)|\\
			&\forall i\in I_k^{\leq}, j\in I_k^+.
		\end{align*}

		Once $I_k^{\leq}$ and $I_k^{+}$ are selected,  we need to verify that
		\begin{equation}\label{pf1}
			|\boldsymbol x^*(i)|<|\boldsymbol x^*(j)|,  \  \forall i\in I^{\leq}_k, j \in I_k^+,
		\end{equation}
		meaning that the index sets $I_k^{\leq}$ and $I_k^{+}$ maintain consistency, so that $\boldsymbol x^*$ satisfies the first-order condition and the local optimality condition \cite{Huang2015}.
		
		To verify  (\ref{pf1}), we need to consider two cases, i,e., $t_k\geq \lambda$ and $t_k< \lambda$.
		
		First, we consider $t_k\geq \lambda$. For $i\in I_k^{\leq}$, from the definition of the $k$-max soft shrinkage operator $\mathbb{S}^{k,\lambda}_{\kmax}(\boldsymbol x)$, it holds  that
		\begin{align*}\label{triangle_0}
			\boldsymbol x^*(i) = &\boldsymbol u^*(i) - \lambda \sgn(\boldsymbol u^*)(i)),
		\end{align*}
		or equivalently,
		\begin{equation}\label{triangle}
			(\Phi^T(\boldsymbol y-\Phi \boldsymbol x^*))(i) = \lambda \sgn(\boldsymbol u^*)(i)),
		\end{equation}
		which follows that $  |\boldsymbol x^*(i)+\lambda \sgn(\boldsymbol x^*(i)+(\Phi^T(\boldsymbol y-\Phi \boldsymbol x^*))(i))| = |\boldsymbol x^*(i)+(\Phi^T(\boldsymbol y-\Phi \boldsymbol x^*))(i)| \leq t_k$. Then based on the triangle inequality, we have
		\begin{equation}\label{condition1_1}
			|\boldsymbol x^*(i)| \leq t_k + \lambda.
		\end{equation}

		For $j\in I^+_k$,  we have $\boldsymbol x^*(j) = \boldsymbol u^*(j)$ in $\mathbb{S}^{k,\lambda}_{\kmax}(\boldsymbol x)$, such that $|\boldsymbol x^*(j)|  = |\boldsymbol x^*(j) + (\Phi^T(\boldsymbol y-\Phi \boldsymbol x^*))(j)| \geq t_{k}^+$. From  (\ref{eq_localoptimal_add}), we have 
		\begin{equation}\label{condition1_2}
			|\boldsymbol x^*(j)| \geq t_k^+ > t_k + \lambda.
		\end{equation}
		
		With  (\ref{condition1_1}) and (\ref{condition1_2}), we then have $|\boldsymbol x^*(i)| <  |\boldsymbol x^*(j)|,  \  \forall i\in I^{\leq}_k, j \in I_k^+.$

		Similarly, we can analyze the case with $t_k < \lambda$. From (\ref{soft_new}), it is obvious that $|\boldsymbol x^*(i)|=0$ for $i\in I_k^{\leq}$.  For $j\in I^+_k$,  we have  $|\boldsymbol x^*(j) | = |\boldsymbol u^*(j)|  >  |\boldsymbol x^*(i)|=0$ for $i\in I_k^{\leq}$, implying that  (\ref{pf1}) also holds for $t_k < \lambda$. Therefore, we have $|\boldsymbol x^*(i)|<|\boldsymbol x^*(j)|,  \  \forall i\in I^{\leq}_k, j \in I_k^+$.
		
		We can define a subregion $D_{I_k^{\leq}}\triangleq \{\boldsymbol x: |\boldsymbol x(i)|<|\boldsymbol x(j)|,   \forall i\in I^{\leq}_k, j \in I_k^+\}$, and there is
		\begin{equation}\label{domain}
			\boldsymbol x^*\in D_{I_k}^{\leq}, {U}(\boldsymbol x^*,\delta)\subset D_{I_k^{\leq}}.
		\end{equation}
	
		Then, we can conclude with the following facts, i.e.,
		\begin{itemize}
			\item if  (\ref{eq_localoptimal}) and (\ref{eq_localoptimal_add}) hold for given $I_k^{\leq}$, then $\boldsymbol x^*$ is the minimum of
				\begin{align*}
					\frac{1}{2}||\Phi \boldsymbol x- \boldsymbol y ||_{2}^2 + \lambda \sum_{i\in I_k^{\leq}}|\boldsymbol x(i)|,
				\end{align*}
			\item $L_{\kmax}^{k,\lambda}(\boldsymbol x) = \sum_{I_k^{\leq}}|\boldsymbol x(i)|, \forall \boldsymbol x\in D_{I_k^{\leq}} $.
		\end{itemize}
		In summary,  (\ref{eq_localoptimal}) and (\ref{eq_localoptimal_add})  for $I_k^{\leq}$ guarantee that $\boldsymbol x^*$ is the minimum of $\frac{1}{2}||\Phi \boldsymbol x- \boldsymbol y ||_2^2+\lambda L_{\kmax}^{k,\lambda}(\boldsymbol x)$, $\forall \boldsymbol x \in D_{I_k}^{\leq}$.  Together with (\ref{domain}), we know that $\forall x\in {U}(\boldsymbol x^*,\delta)$,
		\begin{align*}
			&\frac{1}{2}||\Phi \boldsymbol x^*- \boldsymbol y ||_{2}^2+\lambda L_{\kmax}^{k, \lambda}(\boldsymbol x^*)  \\
            & \leq \frac{1}{2}||\Phi\boldsymbol  x- \boldsymbol y ||_{2}^2+\lambda L_{\kmax}^{k,\lambda}(\boldsymbol x).
		\end{align*}
	\end{proof}

    Hence, we can obtain the following theorem  which gives the local optimality condition for the sparse group $k$-max regularization problem (\ref{prob:group_kmax_optimization}).
	\begin{theorem}\label{localoptimal_group}
		If  $\exists  \boldsymbol x^* = [(\boldsymbol{x}_1^*)^T, \ldots, (\boldsymbol{x}_m^*)^T]^T \in \mathbb R^d$,  for all possible $I_{k_i}^{\leq}(\boldsymbol{u}_i^*)$, where $\boldsymbol{u}_i^* = \boldsymbol{x}_i^*+ \Phi_i^T(\boldsymbol y- \sum_{i=1}^m \Phi_i \boldsymbol{x}_i^*)$, such that
		\begin{align}
			&\boldsymbol{x}_i^* = \mathbb{S}_{\grpkmax}^{k_i, \lambda}(\boldsymbol{u}_i^*)\label{eq_localoptimal_group},\\
			&t_{k_i}^+ >t_{k_i} + \lambda, \label{eq_localoptimal_group_add}
		\end{align}
		where  $t_{k_i}$ is the $k_i$-th maximal absolute entry of   $\boldsymbol{u}_i^*$, and $t_{k_i}^+ = \min_{j_i}\{\boldsymbol{u}_i^*(j_i)\}$ with ${j_i\in I_{k_i}^+(\boldsymbol{u}_i^*)}$, then $\boldsymbol x^*$ is a local optimum to   (\ref{prob:group_kmax_optimization}).
	\end{theorem}

	\begin{proof}
		Based on the proof of Theorem \ref{localoptimal}, the index set $I_{k}^{\leq}(\boldsymbol x)$ in  (\ref{eq_stationary_group}) is verified to be consistent, and thus the local optimality in Theorem \ref{localoptimal} holds true for  (\ref{prob:group_kmax_optimization_simple}), since  (\ref{eq_stationary_group}) is the first-order condition for a fixed  $I_{k}^{\leq}(\boldsymbol x)$ in  (\ref{prob:group_kmax_optimization_simple}) \cite{Huang2015}.
		
		Equivalently, the local optimality conditions (\ref{eq_localoptimal_group}) and (\ref{eq_localoptimal_group_add}) also hold true for  (\ref{prob:group_kmax_optimization}) when   $I_{k_i}^{\leq}(\boldsymbol x_i)$ in (\ref{eq_stationary_group}) is verified to be consistent, $ \forall i\in [m]$.
		
		In fact, $\boldsymbol{x}_i$ ($\forall i\in [m]$) are independent to each other, since the second term in (\ref{prob:group_kmax_optimization}) is separable with respect to $\boldsymbol{x}_i$. It results in that  $I_{k_i}^{\leq}(\boldsymbol{x}_i)$  $\forall i \in [m]$ are independent. Combined with Theorem \ref{localoptimal}, we can  proves that  $I_{k_i}^{\leq}(\boldsymbol{x}_i)$ can maintain consistency $\forall i\in [m]$ analogously. Hence, we can verify that $\boldsymbol x^*$ is locally optimal to   (\ref{prob:group_kmax_optimization}).
	\end{proof}

\subsection{Complexity Analysis}
    In a single iteration of Algorithm \ref{alg:grpkmax}, the computational consumption refers to the calculation of the shrinkage operator, especially the calculation of  $\boldsymbol{x}^{(t)}_i + \Phi_i^T(\boldsymbol y - \sum_{i=1}^{m}\Phi_i \boldsymbol{x}^{(t)}_i)$ and the determination of the index sets $I_{k_i}^{\leq}(\boldsymbol{x}_i^{(t)})$, $\forall i \in [m]$. It is obvious that the former term requires only basic operations, such as vector addition and the multiplication of matrices and vectors, while the latter requires the partition of the entries of a vector. More specifically, the time complexity for each $i \in [m]$ is $O(nd_i) + O(d_i + k_i \log k_i)$, which results in the total time complexity in one single iteration being $O(nd + \sum_{i \in [m]}k_i \log k_i)$.

\section{NUMERICAL EXPERIMENTS}\label{sec:exp}

	In this section, we conduct numerical experiments to evaluate the performance of the proposed sparse group $k$-max regularized problem with the corresponding IST algorithm, compared to several basline methods. The experiments are conducted at two levels. First, we evaluate on synthetic data to demonstrate the performance especially on the group-wise and in-group sparsity with prior information. Then, experiments on public datasets are further conducted. All experiments are performed on Matlab R2017b with Intel i7-4790, 3.60GHz CPU, 16G RAM on Windows 10.
	
\subsection{Synthetic  Examples}\label{sec:exp:synthetic_sec}

    The synthetic dataset is generated following \cite{Simon2013a}. In details, we set $n = 200$, $m = 10$, $d_i = 10$, $\forall i \in [m]$, for  (\ref{prob:l0-min}). We first generate the matrices $\Phi_i \in \mathbb{R}^{n \times d_i}$ randomly with each column following the independent identically distributed (i.i.d.) Gaussian distribution. Letting $\boldsymbol s= [s_1, \ldots, s_m]^T=[10, 8, 6, 4, 2, 1, 0, 0, 0, 0]^T$, we generate $\boldsymbol{x}_i \in \mathbb{R}^{d_i}$ by fixing its randomly selected $s_i$ entries as  $-1$ or $1$, and letting the others be $0$, $\forall i$. We also generate $\boldsymbol{\epsilon} \in \mathbb{R}^{n}$ following i.i.d. Gaussian with the variance $\sigma^2 = 4$. Then, we have 
    \begin{equation}
	\boldsymbol y = \sum_{i=1}^m \Phi_i \boldsymbol{x}_i + \boldsymbol \epsilon.
    \end{equation}

    We choose the following models and corresponding algorithms for comparison.
    \begin{enumerate}[label=(\arabic*),itemindent=15pt,labelsep=3pt,fullwidth,parsep=0pt,labelwidth=0pt]
        \item \textit{\lasso} in the form of (\ref{prob:l1-norm}), which is solved via the iterative thresholding algorithm proposed in \cite{Daubechies2004}.

        \item Group lasso (\textit{\glasso}) in the form of (\ref{prob:grplasso}), which is solved in \cite{Liu2010}.

        \item Sparse group lasso (\textit{\sglasso}) in the form of (\ref{prob:spgrplasso}), which is solved in \cite{Liu2010}.

        \item The proposed sparse group $k$-max regularization (\textit{\gkmax (P)}) in the form of (\ref{prob:group_kmax_optimization})  solved via Algorithm \ref{alg:grpkmax}, with $k_i$ ($\forall i$) known in advance.

        \item The proposed regularization (\textit{\gkmax}) with $k_i$ initialized via the results obtained by \textit{\lasso}.
    \end{enumerate}

    The common settings of these methods are as follows. Determine the regularization parameter $\lambda$  from $\{ 1e-3, \cdots, 1e-2 \}$ and $\mu$ from $\{ 0.5, 0.6, \ldots, 1\}$  via 10-fold cross-validation. The grouping strategies remain the same. We set the maximal iteration number $T=500$ and the termination precision $\delta = 1e{-4}$. 
    
\begin{figure}[!htp]
        \centering
        \begin{subfigure}[b]{0.48\columnwidth}
             \centering
             \includegraphics[width=\columnwidth]{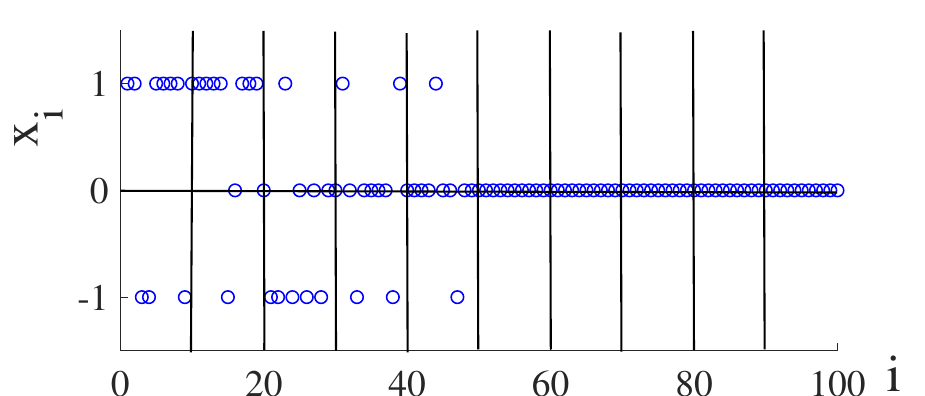}
             \caption{True signals}
             \label{fig_true}
        \end{subfigure}\\
        \hfill
        \begin{subfigure}[b]{0.48\columnwidth}
             \centering
             \includegraphics[width=\columnwidth]{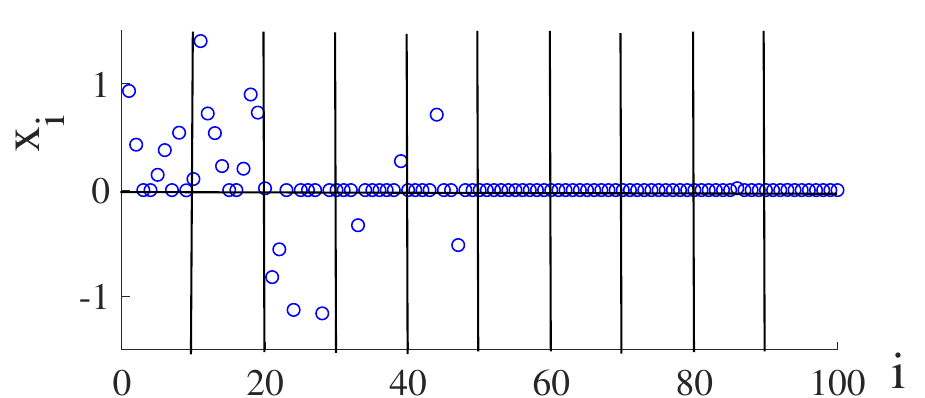}
             \caption{\textit{\lasso}}
             \label{fig_lasso}
        \end{subfigure}
        \begin{subfigure}[b]{0.48\columnwidth}
             \centering
             \includegraphics[width=\columnwidth]{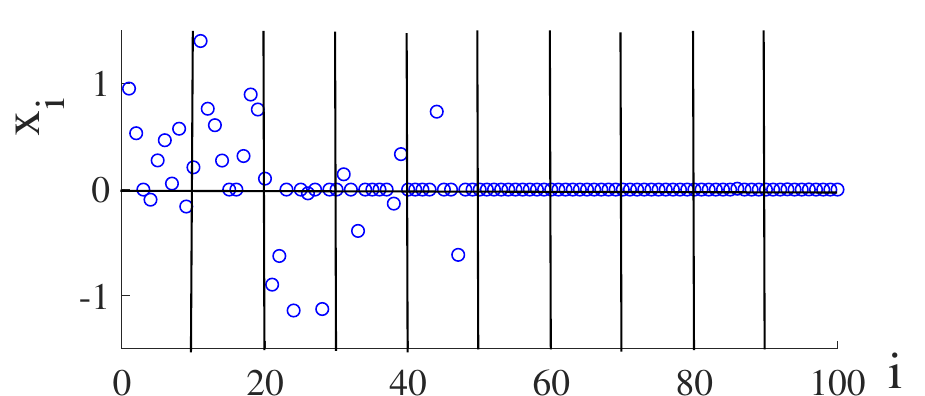}
             \caption{\textit{\glasso}}
             \label{fig_grp_lasso}
        \end{subfigure}
        \begin{subfigure}[b]{0.48\columnwidth}
             \centering
             \includegraphics[width=\columnwidth]{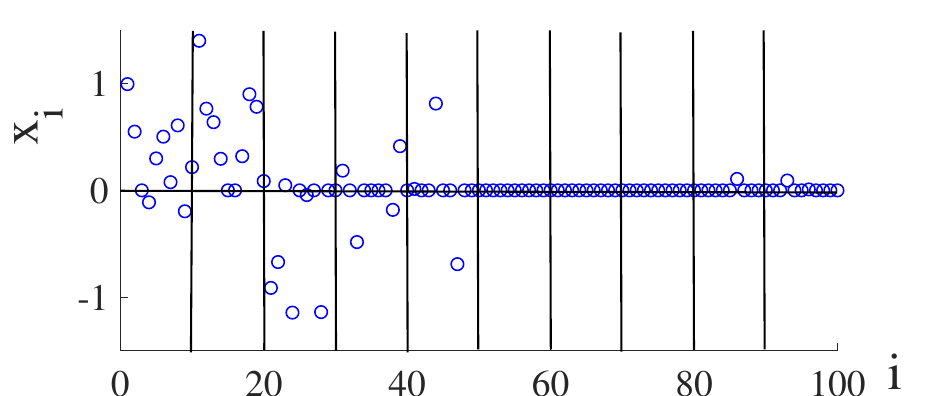}
             \caption{\textit{\sglasso}}
             \label{fig_grp_sp}
        \end{subfigure}
		\begin{subfigure}[b]{0.48\columnwidth}
             \centering
             \includegraphics[width=\columnwidth]{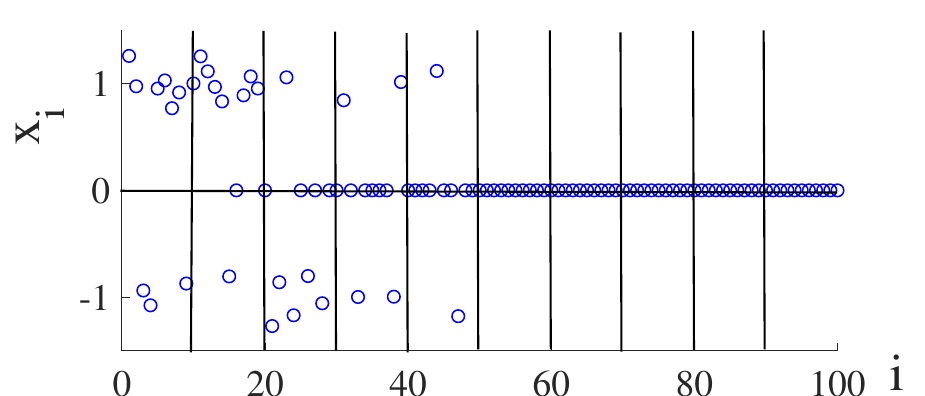}
             \caption{\textit{\gkmax (P)}}
             \label{fig_grp_kmax}
        \end{subfigure}
		\caption{\small Visualization on the recovery of sparse signals with grouping characteristics using different regularizations. The horizontal axis denotes the index of entries in each $\boldsymbol{x}_i$ in the vector $\boldsymbol{x} = [\boldsymbol{x}_1^T, \cdots, \boldsymbol{x}_m^T]^T$, and the vertical axis represents the corresponding signal (variable) values. }\label{toy_signal_result}
	\end{figure}

    We first run experiments on \textit{\lasso}, \textit{\glasso}, \textit{\sglasso}, and \textit{\gkmax (P)} to compare the performance on recovering sparse signals, and the results are presented in Fig. \ref{toy_signal_result}. We observe that with prior information on the in-group sparsity, i.e., $k_i$, the proposed regularization can not only achieve more sparse predictions, but also result in more accurate recover of non-zero variables with respect to their magnitudes. In this experiment, we also record the total computing time of each method, i.e., $t_{\rm {\lasso}} =0.18~ \rm{s}$, $t_{\rm {\glasso}}=0.51 ~\rm{s}$, $t_{\rm \sglasso}=0.38~ \rm{s}$ and $t_{k\rm{-MAX (P)}}= 1.15 ~\rm{s}$. The reason for the more computation time lies in the fact that the proposed regularization is non-convex. However, with the improved accuracy, the efficiency is still acceptable for the given scale of problems.

	Since  the ground-truth group-wise/in-group sparsity is usually unknown in real-world applications, we run experiments to empirically analyze how to initialize the hyperparameters, $k_i$, $\forall i$. More specifically, We compare the method, \textit{\gkmax}, which initializes $k_i$ using the results obtained by the lasso, with other methods. To further evaluate the cases without such priori, we vary the number of variable groups, i.e., $m \in \{5, 10, 15, 20 \}$, with $d_i=10m$, $\forall i$. We repeat the tests 20 times and Table \ref{toy_performance} records the average performance, including the root mean squared error (RMSE)  and the correctly predicted ratio (CPR), which denotes the proportion of correct nonzero coefficients. We  mark the best performance  in bold.
    From the results, we observe that the proposed sparse group $k$-max regularization can achieve higher accuracy than the others.  We also find that the proposed sparse group $k$-max regularization maintains advantageous over the baseline methods with groups/dimensionality increasing.

    \begin{table}[!htb]
    \caption{\small  Performances on the synthetic dataset}\label{toy_performance}
    \setlength{\tabcolsep}{1.5mm}{
    \begin{tabular}{ccccc|cccc}
    \toprule
    \multicolumn{1}{l|}{}          & \multicolumn{4}{c|}{CPR (\%)}            & \multicolumn{4}{c}{RMSE (\%)}                                                                            \\ \hline
    \multicolumn{1}{c|}{$m$}      & $5$      & $10$   & $15$     & $20$     & \multicolumn{1}{c}{$5$} & \multicolumn{1}{c}{$10$} & \multicolumn{1}{c}{$15$} & \multicolumn{1}{c}{$20$} \\\midrule
    \multicolumn{1}{c|}{\textit{\lasso}}    & 93.3     & 88.9   & 85.7     & 86.7     & 64.6                    & 65.5                     & 71.2                     & 78.0                     \\
    \multicolumn{1}{c|}{\textit{\glasso}}  & 93.3     & 87.1   & 85.7     & 82.8     & 28.7                    & 25.3                     & 27.6                     & 30.2                     \\
    \multicolumn{1}{c|}{\textit{\sglasso}} & 93.3     & 90.0   & 85.2     & 86.7     & 24.6                    & 25.8                     & 28.1                     & 29.5                     \\
    \multicolumn{1}{c|}{$K$-MAX}  & \textbf{96.7} & \textbf{93.3} & \textbf{89.7} & \textbf{90.0} & \textbf{13.7}                    & \textbf{11.3   }                  & \textbf{17.6}                     & \textbf{23.1}                     \\ \bottomrule
    \end{tabular}}
    \end{table}

\subsection{Diabetes Dataset}\label{sec:exp:diabetes}

	We conduct experiments on a real world dataset, namely the Diabetes dataset \cite{Efron2004}. The dataset contains 10 features to quantitatively measure the disease progression one year after the baseline. The features include the age, sex, body-mass index, average blood pressure, and 6 blood serum measurements, and there are totally 442 samples, each of which corresponds to the observations of a single patient.

    We divide the 10 features 
    into 3 groups, the first of which consists of the age and sex features, the second the body-mass index and average blood pressure features, and the third the 6 blood serum measurement features.

\begin{figure}[!htp]
        \begin{subfigure}[b]{0.45\columnwidth}
             \centering
             \includegraphics[width=\columnwidth]{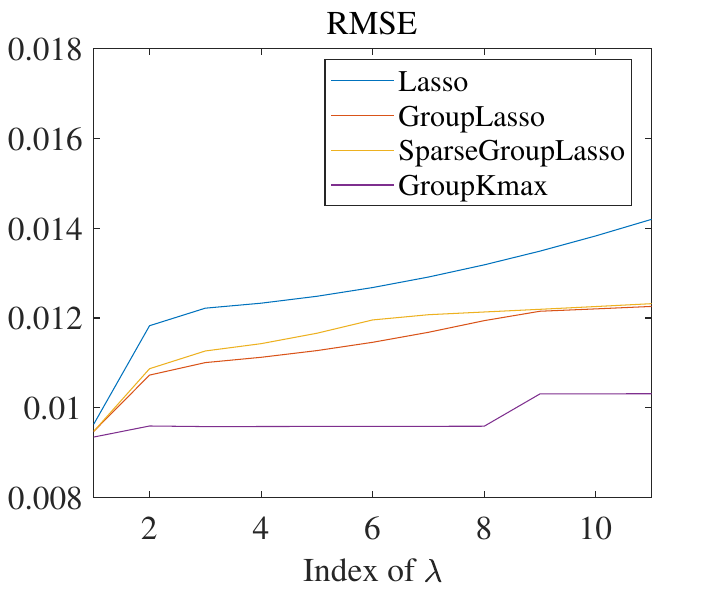}
             \caption{\small RMSE}
             \label{fig_diabetes_rsse}
        \end{subfigure}
		\begin{subfigure}[b]{0.45\columnwidth}
             \centering
             \includegraphics[width=\columnwidth]{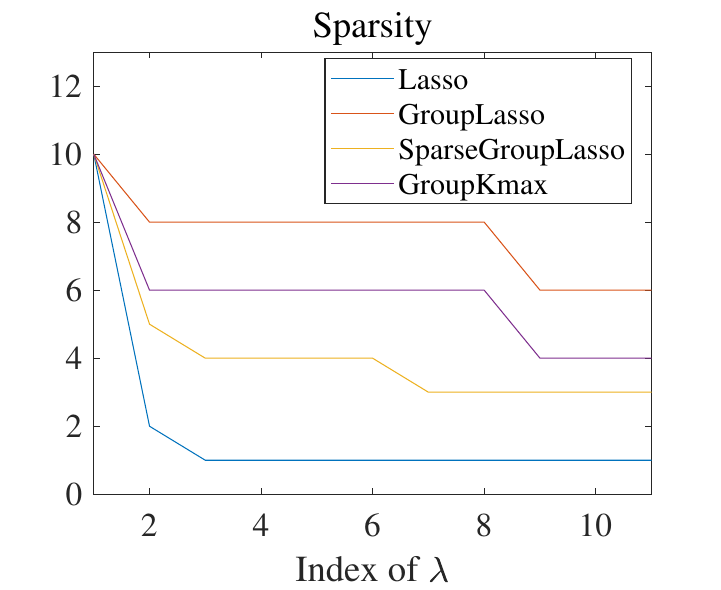}
             \caption{\small Sparsity}
             \label{fig_diabetes_sparsity}
        \end{subfigure}
		\caption{\small  Performance on the diabetes dataset. }\label{diabetes_result}
	\end{figure}

	We compare \textit{\lasso}, \textit{\glasso}, \textit{\sglasso}, and \textit{\gkmax} with   the same settings as in Section \ref{sec:exp:synthetic_sec}, and present  the regularization path of different methods in Fig. \ref{diabetes_result}. The total computation time  is recorded as  $t_{\rm{\lasso}} =0.04~\rm{s}$, $t_{\rm {\glasso}}=0.04~\rm{s}$, $t_{\rm{\sglasso}}=0.07~\rm{s}$ and $t_{k\rm{-MAX}}=0.15~\rm{s}$. We can see that the prediction becomes sparser with $\lambda$ increasing and \gkmax can maintain lower RMSE compared to other methods, which verifies its effectiveness. Based on the results of RMSE,  variables including the body-mass index, average blood pressure, and another 4 blood serum measurements, are selected, which can provide a simple reference for variable selection and analysis.

\subsection{Alcoholic EEG Dataset}\label{sec:exp:eeg}
    The Alcoholic EEG dataset \cite{Zhu2014a} is constructed to classify whether a subject has genetic predisposition to alcoholism, based on $16384$-dimension data, which contains measurements from 64 electrodes placed on different location of the scalps which are sampled at 256 Hz. The dataset contains EEG data for 10 alcoholic and 10 control subjects, with 10 runs per subject per paradigm.

    We compare \textit{\glasso}, \textit{\sglasso}, and \textit{\gkmax} in this experiment, and exclude Lasso since it appears distinctive inferior performance as a result of grouped physical motivation of the EEG signals.  We determine  $\lambda \in \{0, 0.01, 0.02, \ldots, 0.2\}$ by 10-fold cross validation, and keep the other settings  the same as in Section \ref{sec:exp:synthetic_sec}. We use the classification accuracy (\textit{ACC.}), the overall sparsity (\textit{O-SPARS.}), and the number of groups (\textit{\#Group}) to evaluate the performance. We report the results in Table \ref{eeg_accuracy}, and observe that both \textit{\glasso} and \textit{\sglasso} yield solutions with no zero entries, i.e., the information obtained from the sampled 64 electrodes is all kept. \textit{\gkmax}  achieves better performance than the other two  with more sparsity, where   information of only $\hat{L}=46$ electrodes is kept.

	
	\begin{table}[!htb]
		\begin{center}
			\caption{\small Performance on the Alcoholic EEG dataset.}\label{eeg_accuracy}
			\begin{tabular}{c|ccccc}
				\hline
				 &  \textit{ACC. (\%)} & \textit{O-SPARS.} &  \textit{\#Group} \\
				\hline
				\textit{\glasso} & 75.17  & 16384&64 \\
				\textit{\sglasso} & 75.17 & 16384&64 \\
				\gkmax & \textbf{75.33}  & \textbf{9614}&\textbf{46} \\
				 \hline
			\end{tabular}
		\end{center}
	\end{table}
 
    To further evaluate these methods, we set the prior information on the number of non-zero groups as $46$ for all methods, and report the results in Table \ref{eeg_accuracy_1}. We can see  that \textit{\gkmax} can maintain advantages in accuracy when the solution has similar sparsity level with those of group lasso and sparse group lasso, which further reflects the effectiveness of the proposed method.

	\begin{table}[!htb]
		\begin{center}
			\caption{\small Performance on the Alcoholic EEG dataset with fixed number of non-sparse groups ($46$).}\label{eeg_accuracy_1}
			\begin{tabular}{c|ccccc}
				\hline
				 &  \textit{\glasso}  & \textit{\sglasso} & \gkmax\\
				\hline
				ACC. (\%) & 72.50 & 72.83 & \textbf{75.33} \\\hline
			\end{tabular}
		\end{center}
	\end{table}


\section{CONCLUSIONS AND FUTURE WORK}

	In the literature, the linear inverse problems with sparsity constraints have been investigated for decades, with numerous models and corresponding optimization algorithms being proposed. 
    However, to the best of our knowledge, there exists few related work that successfully address the challenges to achieve both group-wise and in-group sparsity for certain applications or to restrain the magnitudes only on selected variables. In this paper, we introduce a novel and concise approximation to the $l_0$ norm, i.e., the sparse group $k$-max regularization, which forces the addition of the smallest $(d-k)$ entries of a variable to approach zero. In this way, it can simultaneously enhance the sparsity and alleviate the restraints on the nonzero entries. In addition, the proposed regularization can be easily generalized to the problems with grouped variables, since its formulation is separable, and thus unifies the group-wise sparsity and in-group sparsity together. To solve the corresponding non-convex and non-differentiable optimization problem, we propose a modified soft thresholding algorithm and provide the local optimality conditions and complexity analysis. Numerical experiments verify the effectiveness and advantages of the proposed sparse group $k$-max regularization.
	
	Although the proposed sparse group $k$-max regularization can promote enhanced sparsity into the grouped variables, prior knowledge on the ground-truth in-group sparsity needs to be ready in advance or to be initialized with careful designs. Besides, the current algorithm with the  index-varied objective can also require extra computation to converge.   In future, we can optimize the initialization  of such prior and extend the application with more generalized frameworks and efficient algorithms.
%
%

\bibliographystyle{IEEEtran}
\bibliography{REFERENCES}

%
%
%
%
%

\end{document}